\documentclass{copernicus2_final}

\usepackage{amsmath}
\usepackage{multirow}
\usepackage{listings}
\usepackage{tabu}

\begin{document}

\title{{Recognizing Birds from Sound -\\The 2018 BirdCLEF Baseline System}}

\author[1]{Stefan Kahl}
\author[1]{Thomas Wilhelm-Stein}
\author[3]{Holger Klinck}
\author[2]{Danny Kowerko}
\author[1]{Maximilian Eibl}

\affil[1]{Chair Media Informatics, Chemnitz University of Technology, D-09107 Chemnitz, Germany}
\affil[2]{Junior Professorship Media Computing, Chemnitz University of Technology, D-09107 Chemnitz, Germany}
\affil[3]{Bioacoustics Research Program, Cornell Lab of Ornithology, Cornell University, 159 Sapsucker Woods Road, Ithaca, NY 14850, USA}

\runningtitle{{The 2018 BirdCLEF Baseline System}}
\runningauthor{{Kahl et. al.}}
\correspondence{{Stefan Kahl\\stefan.kahl@informatik.tu-chemnitz.de}}

\received{}
\pubdiscuss{} 
\revised{}
\accepted{}
\published{}


\pagenumbering{arabic}

\firstpage{1}

\abstract{
Reliable identification of bird species in recorded audio files would be a transformative tool for researchers, conservation biologists, and birders. In recent years, artificial neural networks have greatly improved the detection quality of machine learning systems for bird species recognition. We present a baseline system using convolutional neural networks. We publish our code base as reference for participants in the 2018 LifeCLEF bird identification task and discuss our experiments and potential improvements.\\\\The repository and a continuative tutorial can be found here: \textbf{https://github.com/kahst/BirdCLEF-Baseline}
}

\maketitle  

\section{Motivation}\label{motivation}

Birds are meaningful to a wide audience including the public. They live in almost every type of environment and in almost every niche (place or role) within those environments. The monitoring of species diversity and migration is important for almost all conservation efforts. The analysis of long-term audio data is vital to support those efforts but relies on complex algorithms that need to adapt to changing environmental conditions. Recent advances in the field of Deep Learning showed promising results and we attempt to present a baseline for large-scale bird sound recognition using convolutional neural networks (CNN).

\section{Dataset}\label{dataset}

The 2018 LifeCLEF bird identification task \citep{joly2017lifeclef}, \citep{goeau:hal-01629175} features two main sources of audio recordings. The training dataset contains 36,493 monophonic recordings from South America covering 1500 bird species. The recordings originate from xeno-canto, most audio files are sampled at 44.1 kHz, show a wide variety of recording quality and background noise. The training set has a massive class imbalance and is complemented with textual metadata such as foreground and background species, user quality ratings, time and location of the recording and author name and notes.

The test dataset consists of 12,347 monophonic recordings (also taken from xeno-canto) and 6.5h of annotated soundscapes with time-coded labels recorded in Columbia and Peru. 

\section{Workflow}\label{workflow}

Our proposed workflow consists of three main phases: First, we extract spectrograms from audio recordings. Secondly, we train a deep neural net based on the resulting spectrograms - we treat the audio classification task as an image-processing problem. Finally, we test the trained net given a local validation set of unseen audio recordings.

\subsection{Spectrogram Extraction}

Finding the most efficient strategy for spectrogram extraction from audio recordings is a challenging task. MEL-scale log-amplitude spectrograms are a commonly used technique for sound classification in recent years, e.g. \citep{grill2017two}. We are using signal chunks of one-second length, which seems to be a good fit for birdcalls and even most songs. Most bird species vocalize between 0.5 kHz and 10 kHz - some of them use up to 12 kHz of the frequency spectrum \citep{marler2004nature}. We use a high-pass and a low-pass filter with cut-off frequencies of 300 Hz and 15 kHz and safely include the most important parts of all calls and songs.

\begin{figure*}[t]
    \includegraphics[width=1.0\textwidth]{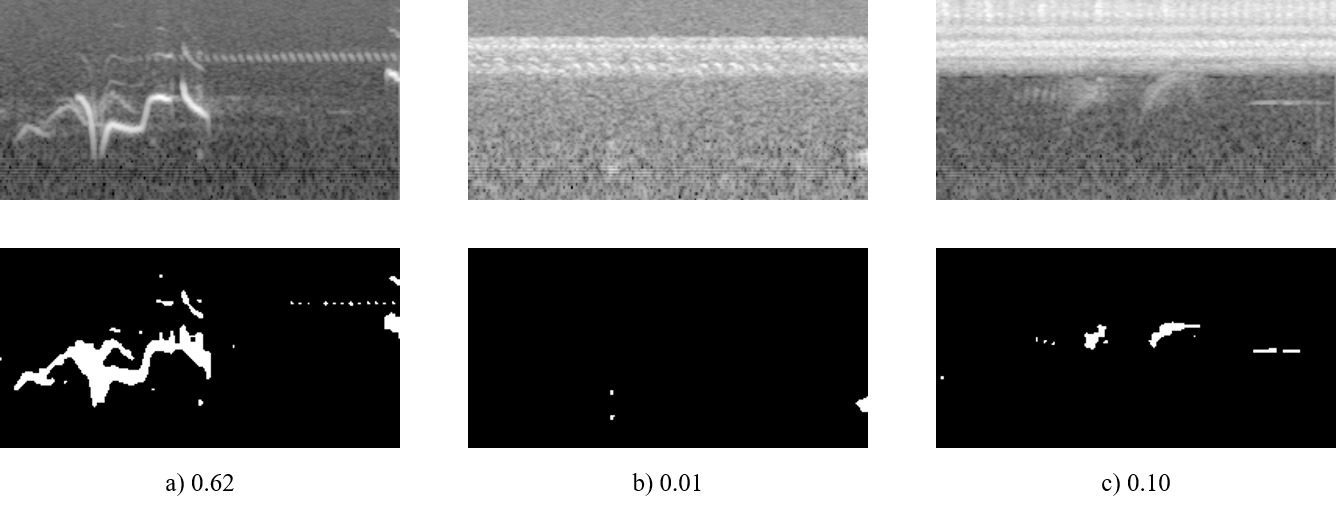}
    \caption{\label{fig:bird_no_bird} Rule-based analysis of extracted spectrograms in search of sufficient signal that qualifies as training sample. a) Clear signal components result in high value, b) steady noise results in low value and can be ignored, c) steady noise in combination with actual bird sound results in higher value and can be accepted as sample.}
    \vspace*{-0.4cm}
\end{figure*}

Additionally, we apply a simple rule-based signal-to-noise ratio estimation based on the attempts of \citep{sprengel2016audio} and \citep{kahl2017large}. Figure~\ref{fig:bird_no_bird} depicts some example spectrograms and the corresponding values which are used to reject samples which (with a high probability) do not contain any bird sounds. This process is not perfect, but fast and results in a good overall ranking of training samples.

\subsection{Training}

We train our baseline model on 521,873 spectrograms extracted from the training data. We use a 5\% validation split to monitor the training process and apply early stopping if the validation error does not progress. We apply a cosine dynamic learning rate schedule mentioned in \citep{huang2017snapshot} starting at 0.001. Despite the adaptive nature of the ADAM optimizer \citep{kingma2014adam}, refining the learning rate after each epoch still seems to be beneficial. Section~\ref{evaluation} details some results of the training process. 

\subsection{Testing}

After the training process completes, we save a snapshot of the best performing model and use that model to evaluate the performance on unseen audio recordings. Again, we split each recording into one-second chunks and predict the most probable class for each spectrogram. Finally, we apply mean exponential pooling to retrieve a single prediction (P) for each class (c) for the entire recording (n specs):

\begin{align*}
P_c = \frac{1}{n}\sum_{i=1}^{n}(2P_{c_i})^2
\end{align*}

We can increase the overall accuracy for monophonic recordings by 1-3\% with that pooling strategy (compared to normal average pooling). However, this assumption might not hold up for soundscapes recordings as the pooling interval is only a few seconds long in that domain (the mandatory prediction interval for the BirdCLEF Soundscape task is five seconds long).


\section{CNN Architecture}\label{neural_net}


The proposed baseline system is built upon a classic CNN architecture that contains seven weighted layers and no bottleneck or shortcut connections. The current net is fully convolutional and does not contain any densely connected layers, except for the final classification layer. 
\begin{table}[h]
\caption{Baseline CNN architecture with corresponding input and output shapes.}
\vskip4mm
\centering
\begin{tabu} to 0.45\textwidth { X[r] X[l] X[l] }
\tophline
Layer&Input Shape&Output Shape\\
\middlehline
Input&(1, 128, 256)&(1, 128, 256)\\
Conv Group 1&(1, 128, 256)&(64, 64, 128)\\
Conv Group 2&(64, 64, 128)&(128, 32, 64)\\
Conv Group 3&(128, 32, 64)&(256, 16, 32)\\
Conv Group 4&(256, 16, 32)&(512, 8, 16)\\
Conv Group 5&(512, 8, 16)&(1024, 4, 8)\\
1x1 Convolution&(1024, 4, 8)&(2048, 4, 8)\\
Global Pooling&(2048, 4, 8)&(2048, 1)\\
Softmax&(2048, 1)&(1500, 1)\\
\middlehline
Weighted Layers&7&\\
Parameters&11,454,236&\\
\bottomhline
\end{tabu}
\end{table}

Each \textbf{Conv Group} contains a convolutional layer with 3x3 kernels, a batch normalization layer \citep{ioffe2015batch} and activation function - ReLU \citep{nair2010rectified} in our case. We provide an implementation of the entire workflow with easy to use settings to make changes to the CNN layout as convenient as possible. The implementation is based on Theano \citep{2016arXiv160502688short} and Lasagne \citep{lasagne} - we will support other frameworks in the future.\\

Our baseline architecture supports grouped convolutions \citep{ioannou2017deep} that are well suited for data based on time-series. We insert a final 1x1 convolution with identity activation to increase the number of features before global average pooling \citep{iandola2016squeezenet}. Increasing the number of filters in each convolution also increases the overall performance but comes at the cost of significantly longer training times.\\


\section{Evaluation}\label{evaluation}

We use a local validation set which comprises 10\% of the original training data and contains at least one sample for every bird species (Full 1500/4399). This set covers all 1500 species and contains 4399 wav-files. Additionally, we use a subset of the local validation set for hyperparameter evaluation (Subset 250/500) which consist of 250 randomly selected classes and is evaluated using 500 randomly selected test audio recordings of the full local validation set.
\\\\
We use the \textbf{Mean Label Ranking Average Precision} as metric which equals the Mean Reciprocal Rank for single labels and the sample-wise Mean Average Precision for multi-label tasks.


\begin{table}[h]
\caption{Results of hyperparameter search - dataset augmentation has significant impact on the performance. Doubling the number of filters and max pooling instead of strided convolutions lead to better results, but significantly increase the computational costs.}
\vskip4mm
\centering
\begin{tabu} to 0.45\textwidth { X[c] X[c] X[c]}
\tophline
Run Name&Validation Set&MLRAP\\
\middlehline
Simple Model&Subset 250/500&0.451\\
+ Batch Norm&Subset 250/500&0.474\\
+ Augmentation&Subset 250/500&0.679\\
+ Filters x2&Subset 250/500&0.695\\
+ Max Pooling&Subset 250/500&0.720\\
+ Post-Pooling&Subset 250/500&0.731\\
\middlehline
Best Model&Full 1500/4399&0.535\\
Best Ensemble&Full 1500/4399&0.564\\
\bottomhline
\end{tabu}
\end{table}

We apply a vertical shift of 10\% and additional noise samples as dataset augmentation methods. Noise samples are the byproduct of the spectrogram extraction process - spectrograms that were rejected by our heuristic, as they do not contain bird sounds with a high probability. Both methods are very effective, act as strong regularizers and improve the performance by a huge margin. Other regularization methods like dropout did not improve the results.\\

Our best single model achieves a multi-label MLRAP of 0.535 on our full local validation set including background species after 70 epochs. Our best ensemble - composed of seven snapshots during training (epoch 10, 20, 30, 40, 50, 60, 70) - achieves a multi-label MLRAP of 0.564.

\section{Improvements}\label{improvements}

We present a baseline system capable of detecting and classifying bird species based on audio recordings with good overall performance. However, our code base allows easy-to-implement improvements that should yield better results once incorporated. Some of the things we could imagine to have great impact on the performance are:

\begin{itemize}
    \item \textbf{More complex net layouts.} Recent advances in the domain of image recognition resulted in very deep CNN architectures with multiple tens of layers, e.g. DenseNet \citep{huang2017densely} or WideResNet \citep{zagoruyko2016wide}. Not all of them are equally suited for the spectrogram domain but might improve the detection performance if tuned carefully. State-of-the-art implementations of alternative models can be found online\footnote{https://github.com/Lasagne/Recipes/tree/master/modelzoo} and can be incorporated easily into our code base.
    \item \textbf{Fine-tuning the spectrogram extraction}. We experimented with many different configurations for the extraction of spectrograms from audio recordings. Changing parameters like number of MELs, resolution, duration or even entirely different spectral representations may result in images that are more detailed and potentially better results. The work of \citep{wang2017trainable} even suggests that log-amplitudes are inferior to per-channel energy normalization (PCEN).
    \item \textbf{Additional metadata.} Pre-selecting bird species based on environmental conditions like habitat, time-of year and weather can have great impact on the quality of predictions. Community projects such as eBird\footnote{https://ebird.org/explore} provide a vast variety of metadata that is freely available. Additionally, the xeno-canto metadata provided with the dataset can be used to build clusters of bird species which can be used for model ensembles. Fusing model predictions and metadata will eliminate implausible predictions and may reduce the false negative rate.
\end{itemize}

We will provide further improvements of the baseline system and additional implementations of different strategies in the future. We encourage all participants of the BirdCLEF challenge to build upon the provided code base and share the results for future reference.

\enlargethispage{\baselineskip}
\begin{acknowledgements}
The European Union and the European Social Fund for Germany partially funded this research. This work was also partially funded by the German Federal Ministry of
Education and Research in the program of Entrepreneurial Regions InnoProfileTransfer in the project group localizeIT (funding code 03IPT608X). Additionally, Jake (Cornell class of '66) and Sue Holshuh kindly supported this project. We want to thank Jan Schl\"uter (OFAI) for his valuable feedback and kind advice. 
\vspace{-0.2cm}
\end{acknowledgements}

%


\bibliographystyle{copernicus}
\bibliography{literatur}

\addtocounter{figure}{-1}\renewcommand{\thefigure}{\arabic{figure}a}

\end{document}